\documentclass{new_tlp}
\usepackage{times}
\usepackage{newtxmath}
\usepackage{booktabs}

\usepackage{caption}
\usepackage{subcaption}
\usepackage{graphicx}
\usepackage{amsmath}                
\usepackage[linesnumbered,lined,ruled]{algorithm2e}
\usepackage{url}

\newcommand\bcmdtab{\noindent\bgroup\tabcolsep=0pt%
  \begin{tabular}{@{}p{10pc}@{}p{20pc}@{}}}
\newcommand\ecmdtab{\end{tabular}\egroup}

\newcommand{\rif}{\stackrel{\,\,+}{\leftarrow}}

\title[KR for Ad Hoc Teamwork]{Knowledge-based Reasoning and Learning under Partial Observability in Ad Hoc Teamwork}

\author[Hasra Dodampegama and Mohan Sridharan]
       {HASRA DODAMPEGAMA and MOHAN SRIDHARAN\\
       Intelligent Robotics Lab, School of Computer Science, University of Birmingham, UK\\
       \email{hhd968@student.bham.ac.uk, m.sridharan@bham.ac.uk}}

\jdate{May 2023}
\pubyear{2023}
\pagerange{\pageref{firstpage}--\pageref{lastpage}}
\doi{S1471068401001193}

\begin{document}

\label{firstpage}

\maketitle

  \begin{abstract}
    Ad hoc teamwork refers to the problem of enabling an agent to collaborate with teammates without prior coordination. Data-driven methods represent the state of the art in ad hoc teamwork. They use a large labeled dataset of prior observations to model the behavior of other agent \textit{types} and to determine the ad hoc agent's behavior. These methods are computationally expensive, lack transparency, and make it difficult to adapt to previously unseen changes, e.g., in team composition. Our recent work introduced an architecture that determined an ad hoc agent's behavior based on non-monotonic logical reasoning with prior commonsense domain knowledge and predictive models of other agents’ behavior that were learned from limited examples. In this paper, we substantially expand the architecture's capabilities to support: (a) online selection, adaptation, and learning of the models that predict the other agents' behavior; and (b) collaboration with teammates in the presence of partial observability and limited communication. We illustrate and experimentally evaluate the capabilities of our architecture in two simulated multiagent benchmark domains for ad hoc teamwork: Fort Attack and Half Field Offense. We show that the performance of our architecture is comparable or better than state of the art data-driven baselines in both simple and complex scenarios, particularly in the presence of limited training data, partial observability, and changes in team composition.
  \end{abstract}




\section{Introduction}
\label{sec:intro}
\begin{figure}[t]
\begin{center}
\begin{subfigure}{.26\columnwidth}
  \centering
  \includegraphics[height=1.2in]{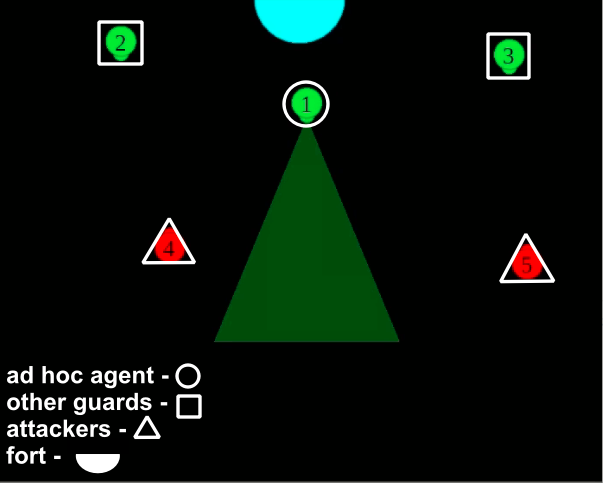}
  \caption{Fully observable}
  \label{fig:fa}
\end{subfigure} \hspace{1em} %
\begin{subfigure}{.26\columnwidth}
  \centering
  \includegraphics[height=1.2in]{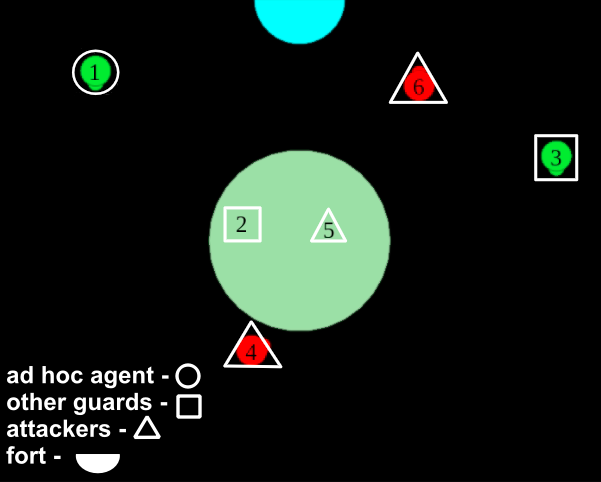}
  \caption{Partially observable}
  \label{fig:fa-l}
\end{subfigure}\hspace{0.5em}
\begin{subfigure}{.22\columnwidth}
  \centering
  \includegraphics[height=1.2in]{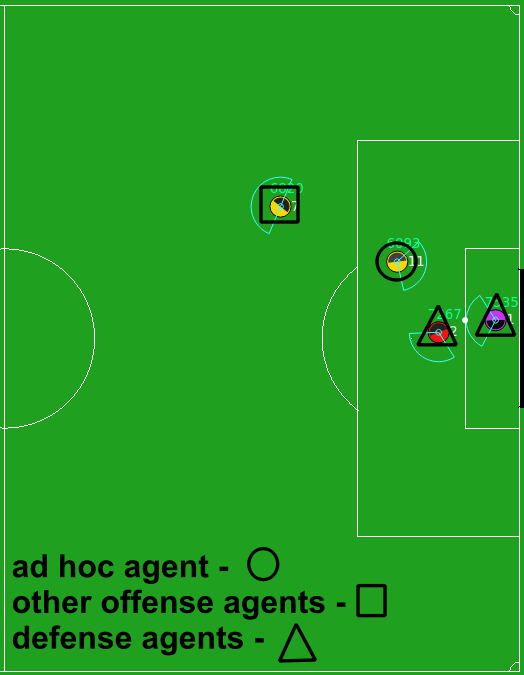}
  \caption{Limited version}
  \label{fig:hfolimited}
\end{subfigure} \hspace{-1.5em}
\begin{subfigure}{.22\columnwidth}
  \centering
  \includegraphics[height=1.2in]{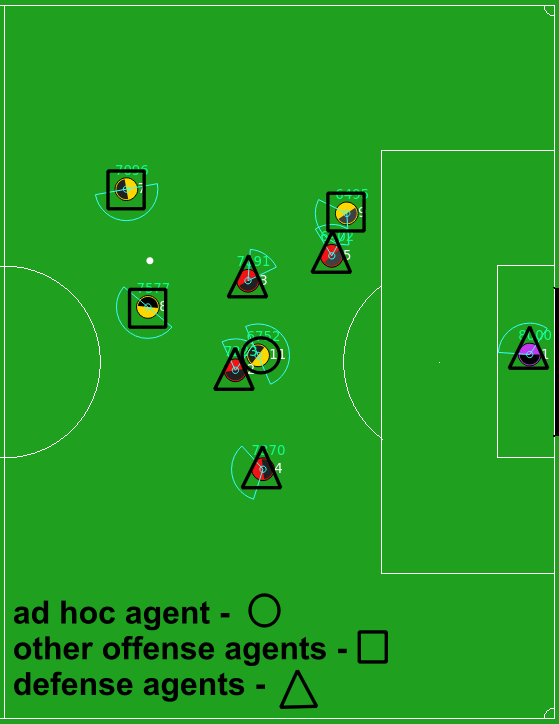}
  \caption{Full version}
  \label{fig:hfofull}
\end{subfigure}
\caption{Screenshots: (a-b) fort attack environment; (c-d) half-field offense environment.}
\label{fig:fa-hfo}
\end{center}
\end{figure}

Ad Hoc Teamwork (AHT) is the challenge of enabling an agent (called the \textit{ad hoc agent}) to collaborate with previously unknown teammates toward a shared goal~\cite{Stone:AAAI10}. As motivating examples, consider the simulated multiagent domain \textit{Fort Attack} (FA, Figure~\ref{fig:fa}), where a team of guards has to protect a fort from a team of attackers~\cite{Deka:20}, and the \textit{Half Field Offense} domain (HFO, Figure~\ref{fig:hfofull}), where a team of offense agents has to score a goal against a team of defenders that includes a goalkeeper~\cite{Hausknecht:ALA16}. Agents in these domains have limited knowledge of each other's capabilities, no prior experience of working as a team, limited ability to observe the environment (Figure~\ref{fig:fa-l}), and limited bandwidth for communication. Such scenarios are representative of multiple practical application domains such as disaster rescue and surveillance. 

The state of the art in AHT has transitioned from the use of predetermined policies for selecting actions in specific states to the use of a key ``data-driven" component. This component uses probabilistic or deep network methods to model the behavior (\emph{i.e.,} action choice in specific states) of other agents or agent types, and to optimize the ad hoc agent's behavior. These methods use a long history of prior experiences in different scenarios, and of the interactions with these agent types, as training examples. It is difficult to obtain such training examples in complex domains, and computationally expensive to build the necessary models or to revise them in response to new situations. At the same time, just reasoning with prior knowledge will not allow the ad hoc agent to accurately anticipate the behavior of other agents and it is not possible to encode comprehensive knowledge about all possible situations. In a departure from existing work, we pursue a \textit{cognitive systems} approach, which recognizes that AHT jointly poses representation, reasoning, and learning challenges, and seeks to leverage the complementary strengths of knowledge-based reasoning and  data-driven learning from limited examples. Specifically, our knowledge-driven AHT architecture (KAT) builds on knowledge representation (KR) tools to support:
\begin{enumerate}
\item Non-monotonic logical reasoning with prior commonsense domain knowledge and rapidly-learned predictive models of other agents' behaviors;
\item Use of reasoning  and observations to trigger the selection and adaptation of relevant agent behavior models, and the learning of new models as needed; and
\item Use of reasoning to guide collaboration with teammates under partial observability.
\end{enumerate}
In this paper, we build on and significantly extend our recent work that demonstrated just the first capability (listed above) in the FA domain~\cite{dodampegama:aaai23}. We use Answer Set Prolog (ASP) for non-monotonic logical reasoning, and heuristic methods based on ecological rationality principles~\cite{gigerenzer:bookchap20} for rapidly learning and revising agents' behavior models. We evaluate KAT's capabilities in the FA domain and the more complex HFO domain. We demonstrate that KAT's performance is better than that of just the non-monotonic logical reasoning component, and is comparable or better than state of the art data-driven methods, particularly in the presence of partial observability and changes in team composition.

\section{Related Work}
\label{sec:relwork}
Methods for AHT have been developed under different names and in different communities over many years, as described in a recent survey~\cite{mirsky:eumas22}. Early work used specific protocols (`plays') to define how an agent should behave in different scenarios (states)~\cite{Bowling:AAAI05}. Subsequent work used sample-based methods such as Upper Confidence bounds for Trees (UCT)~\cite{Barrett:AAAI13}, 
or combined UCT with methods that learned models from historical data for online planning~\cite{Wu:ijcai11}. 
More recent methods have included a key data-driven component, using probabilistic, deep-network, and reinforcement learning (RL)-based methods to learn action (behavior) choice policies for different \textit{types} of agents based on a long history of prior observations of similar agents or situations~\cite{Barrett:AIJ17,Rahman:ICML21}. For example, RL methods have been used to choose the most useful policy (from a set of learned policies) to control the ad hoc agent in each situation~\cite{Barrett:AIJ17}, or to consider predictions from learned policies when selecting an ad hoc agent's actions for different types of agents~\cite{santos:bookchap21}. Attention-based deep neural networks have been used to jointly learn policies for different agent types~\cite{Chen:AAAI20} and for different team compositions~\cite{Rahman:ICML21}. Other work has combined sampling strategies with learning methods to optimize performance~\cite{Zand:IFAAMAS22}.
There has also been work on using deep networks to learn sequential and hierarchical models that are combined with approximate belief inference methods to achieve teamwork under ad hoc settings~\cite{zintgraf:21}. 

Researchers have explored different communication strategies for AHT. Examples include a multiagent, multi-armed bandit formulation to broadcast messages to teammates at a cost~\cite{Barrett:AIJ17}, and a heuristic method to assess the cost and value of different queries to be considered for communication~\cite{Macke:AAAI21}. These methods, similar to the data-driven methods for AHT discussed above, require considerable resources (\emph{e.g.,} computation, training examples), build opaque models, and make it difficult to adapt to changes in team composition. 


There has been considerable research in developing action languages and logics for single- and multiagent domains. This includes action language $\mathcal{A}$ for an agent computing cooperative actions in multiagent domains~\cite{son:bookchap10}, and action language $\mathcal{C}$ for modeling benchmark multiagent domains with minimal extensions~\cite{Baral:Springer10}. Action language $\mathcal{B}$ has also been combined with Prolog and ASP to implement a distributed multiagent planning system that supports communication in a team of collaborative agents~\cite{Son:Springer10}. More recent work has used $\mathcal{B}$ for planning in single agents and multiagent teams, including a distributed approach based on negotiations for non-cooperative or partially-collaborative agents~\cite{Son:KI18}. To model realistic interactions and revise the domain knowledge of agents, researchers have introduced specific action types, \emph{e.g.,} world-altering, sensing, and communication actions~\cite{Baral:AAMAS10}.  
Recent work has represented these action types in action language m$\mathcal{A*}$ while also supporting epistemic planning and dynamic awareness of action occurrences~\cite{Baral:AI22}. These studies have demonstrated the expressive power and reasoning capabilities that logics in general, and non-monotonic logics such as ASP in particular, provide in the context of multiagent systems. Our work draws on these findings to address the reasoning and learning challenges faced by an ad hoc agent that has to collaborate with teammates without any prior coordination, and under conditions of partial observability and limited communication. 

\section{Architecture}
\label{sec:arch}
\begin{figure*}[tb]
\centering
\includegraphics[width=\textwidth]{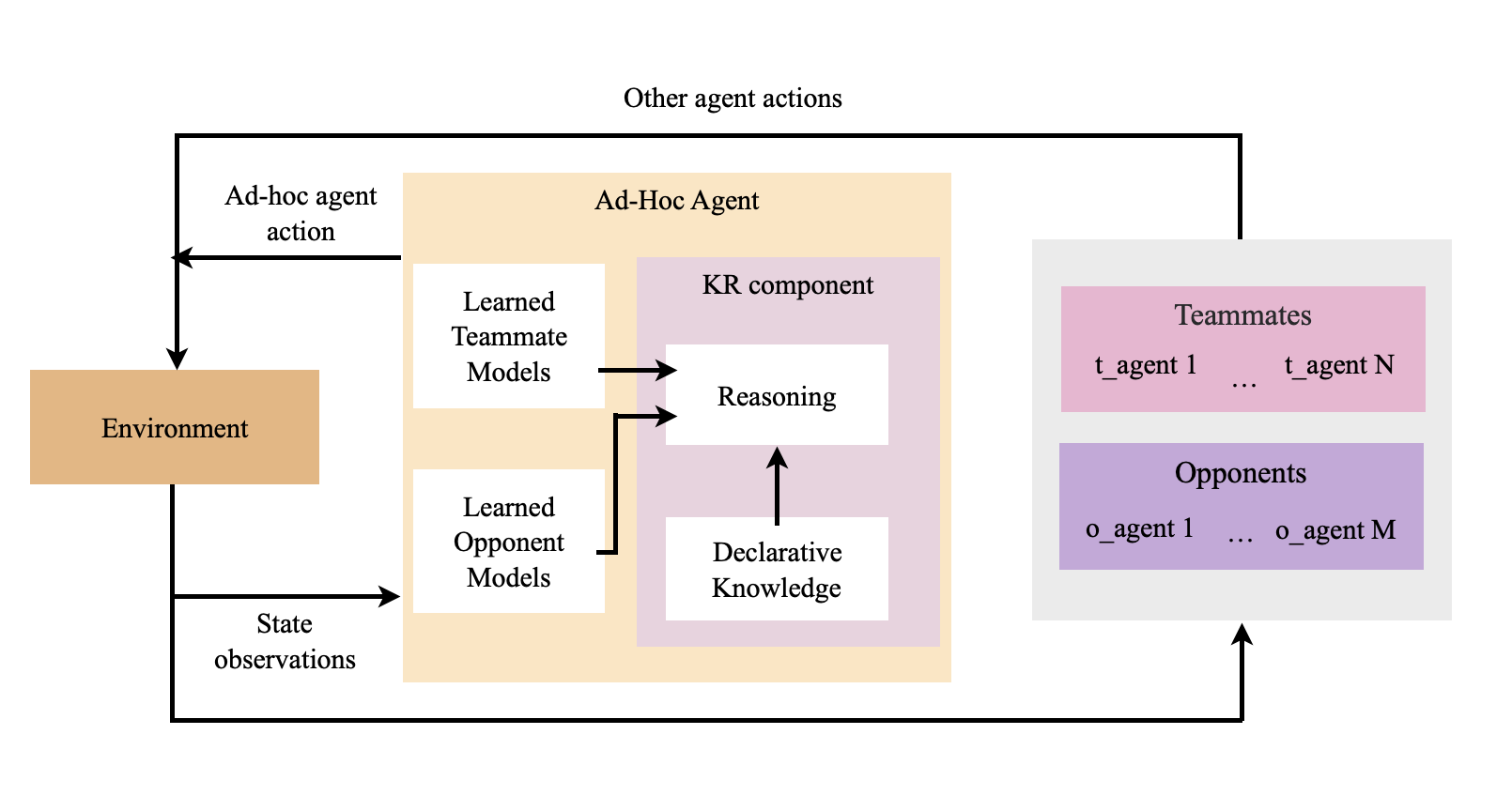}
\vspace{-2.5em}
\caption{Our KAT architecture combines complementary strengths of knowledge-based and data-driven heuristic reasoning and learning.}
\label{fig2}
\end{figure*}

Figure~\ref{fig2} provides an overview of our KAT architecture. KAT enables an ad hoc agent to perform non-monotonic logical reasoning with prior commonsense domain knowledge, and with incrementally learned  behavior models of teammate and opponent agents. At each step, valid observations of the domain state are available to all the agents. Each agent uses these observations to independently determine and execute its individual actions in the domain. The components of KAT are described below using the following two example domains.

\medskip
\noindent
\textbf{Example Domain 1: Fort Attack (FA).} Three guards are protecting a fort from three attackers. One guard is the ad hoc agent that can adapt to changes in the domain and team composition. An episode ends if: (a) guards manage to protect the fort for a period of time; (b) all members of a team are eliminated; or (c) an attacker reaches the fort. 

At each step, each agent can move in one of the four cardinal direction with a particular velocity, turn clockwise or anticlockwise, do nothing, or shoot to kill any agent of the opposing team that is within its shooting range. The environment provides four types of built-in policies for guards and attackers (see Section~\ref{sec:expres-setup}). The original FA domain is fully observable, \emph{i.e.,} each agent knows the state of other agents at each step. We simulate partial observability by creating a ``\textit{forest}" in Figure~\ref{fig:fa-l}; any agent in this region is hidden from others.

\medskip
\noindent
\textbf{Example Domain 2: Half Field Offense (HFO).} This simulated 2D soccer domain is a complex benchmark for multiagent systems and AHT~\cite{Hausknecht:ALA16}. Each game (\textit{i.e.,} episode) is essentially played in one half of the field. The ad hoc agent is one of the members of the offense team that seeks to score a goal against agents in the team defending the goal. 
An episode ends when the: (a) offense team scores a goal; (b) ball leaves field; (c) defense team captures the ball; or (d) maximum episode length (500) is reached.

There are two version of the HFO domain: (i) \textit{limited}: with two offense agents and two defense agents (including the goalkeeper); and (ii) \textit{full}: with four offense agents and five defense agents (including the goalkeeper). Similar to prior AHT methods, agents other than the ad hoc agent are selected from teams created in the RoboCup 2D simulation league competitions. Specifically, other offense team agents are based on the binary files of five teams: \textit{helios, gliders, cyrus, axiom, aut}. For defenders, we use \textit{agent2D} agents, whose policy was derived from \textit{helios}. The strategies of these agent types were trained using data-driven (probabilistic, deep, reinforcement) learning methods. HFO supports two state space abstractions: low and high; we use the high-level features. In addition, there are three abstractions of the action space: primitive, mid-level, and high-level; we use a combination of mid-level and high-level actions. This choice of representation was made to facilitate comparison with existing work.

\smallskip
\noindent
Prior commonsense knowledge in these two domains includes relational descriptions of some domain attributes (\emph{e.g.,} safe regions), agent attributes (\emph{e.g.,} location), default statements, and axioms governing change in the domain, \emph{e.g.,} an agent can only move to a location nearby, only shoot others within its range (FA), and only score a goal from a certain angle (HFO). Specific examples of this knowledge are provided later in this section. Although this knowledge may need to be revised over time in response to changes in the domain, we do not explore knowledge acquisition and revision in this paper; for related work by others in our group, please see the papers by~\citeN{mohan:JAAMAS23} and ~\citeN{mohan:ACS18}. 

\subsection{Knowledge Representation and Reasoning}
\label{sec:arch-krr}
In KAT, the transition diagrams of any domains are described in an extension of the action language $\mathcal{AL}_d$~\cite{gelfond:ANCL13}. Action languages are formal models of parts of natural language that are used to describe the transition diagrams of any given dynamic domain. The domain representation comprises a system description $\mathcal{D}$, a collection of statements of $\mathcal{AL}_d$, and a history $\mathcal{H}$. $\mathcal{D}$ has a sorted signature $\Sigma$ which consists of \emph{actions}, \emph{statics}, \emph{i.e.,} domain attributes whose values cannot be changed, and \emph{fluents}, \emph{i.e.,} domain attributes whose values can be changed by actions. For example, $\Sigma$ in the HFO domain includes basic sorts such as $ad\_hoc\_agent$, $external\_agent$, $agent$, $of\!\!fense\_agent$, $de\!f\!ense\_agent$, $x\_val$, $y\_val$, and sort $step$ for temporal reasoning. Sorts are organized hierarchically, with some sorts, \emph{e.g.,} $of\!\!fense\_agent$ and $de\!f\!ense\_agent$, being subsorts of others, \emph{e.g.,} $external\_agent$. Statics in $\Sigma$ are relations such as $next\_to(x\_val,$ $ y\_val, x\_val, y\_val)$ that encode the relative arrangement of locations (in the HFO domain). The fluents in $\Sigma$ include \emph{inertial} fluents that obey inertia laws and can be changed by actions, and \emph{defined} fluents that do not obey inertia laws and are not changed directly by actions. For example, inertial fluents in the HFO domain include:
\begin{align}
    \label{eqn:inertial-fluent}
   &loc(ad\_hoc\_agent, x\_val, y\_val)\\\nonumber
   &ball\_loc(x\_val, y\_val)\\\nonumber 
   &has\_ball(agent)
\end{align}
which describe the location of the ad hoc agent, location of the ball, and the agent that has control of the ball; the value of these attributes changes as a direct consequence of executing specific actions. Defined fluents of the HFO domain include:
\begin{align}
   \label{eqn:defined-fluent}
   &agent\_loc(external\_agent, x\_val, y\_val) \\ \nonumber
   &de\!f\!ense\_close(agent, defense\_agent) \\\nonumber 
   &far\_from\_goal(ad\_hoc\_agent)
\end{align}
which encode the location of the external (\emph{i.e.,} non-ad hoc) agents, and describe whether a defense agent is too close to another agent, and whether the ad hoc agent is far from the goal. Note that the ad hoc agent has no control over the value of these fluents, although the hoc agent's actions can influence the value of these fluents. Next, actions in the HFO domain include:
\begin{align}
    \label{eqn:action}
&move(ad\_hoc\_agent, x\_val, y\_val)\\\nonumber
&kick\_goal(ad\_hoc\_agent)\\ \nonumber
&dribble(ad\_hoc\_agent, x\_val, y\_val)\\\nonumber
&pass(ad\_hoc\_agent, of\!fense\_agent)
\end{align}
which state the ad hoc agent's ability to move to a location, kick the ball toward the goal, dribble the ball to a location, and pass the ball to a teammate. Next, axioms in $\mathcal{D}$ describe domain dynamics using elements in $\Sigma$ in three types of statements: causal laws, state constraints, and executability conditions. For the HFO domain, this includes statements such as: 
\begin{subequations}
\label{eqn:axioms}
\begin{align}
    &move(R,X,Y) ~\mathbf{ causes }~ loc(R,X,Y) \\
    &dribble(R,X,Y) ~\mathbf{ causes }~ ball\_loc(X,Y)\\
    &\neg has\_ball(A1) ~\mathbf{ if }~ has\_ball(A2), ~A1 \neq A2\\
    &\mathbf{impossible }~ shoot(R) ~\mathbf{ if }~ far\_from\_goal(R)
\end{align}
\end{subequations}
where Statements~\ref{eqn:axioms}(a-b) are causal laws that specify that moving and dribbling change the ad hoc agent's and ball's location (respectively) to the desired location. Statement~\ref{eqn:axioms}(c) is a state constraint that implies that only one agent can control the ball at any time. Statement~\ref{eqn:axioms}(d) is an executability condition that prevents the consideration of a shooting action (during planning) if the ad hoc agent is far from the goal. Finally, the history $\mathcal{H}$ is a record of observations of fluents at particular time steps, \emph{i.e.,} $obs(fluent, boolean, step)$, and of action execution at particular time steps, \emph{i.e.,} $hpd(action, step)$. It also includes initial state defaults, \emph{i.e.,} statements in the initial state that are believed to be true in all but a few exceptional circumstances, \emph{e.g.,} the following $\mathcal{AL}_d$ statement implies that attackers in the FA domain usually spread and attack the fort:
\begin{align}
    \mathbf{initial~~default} ~spread\_attack(X)~ \textbf{if}~~ attacker(X)
\end{align}
To enable an ad hoc agent to reason with prior knowledge, the domain description in $\mathcal{AL}_d$ is automatically translated to program $\Pi(\mathcal{D}, \mathcal{H})$ in CR-Prolog~\cite{balduccini:aaaisymp03}, an extension to ASP that supports consistency restoring (CR) rules. ASP is based on stable model semantics and represents constructs difficult to express in classical logic formalisms. It encodes concepts such as \emph{default negation} and \emph{epistemic disjunction}, and supports non-monotonic reasoning; this ability to revise previously held conclusions is essential in AHT. $\Pi(\mathcal{D}, \mathcal{H})$ incorporates the relation $holds(fluent, step)$ to state that a particular fluent is true at a given step, and $occurs(action, step)$ to state that a particular action occurs in a plan at a given step.  It includes the signature and axioms of $\mathcal{D}$, inertia axioms, reality checks, closed world assumptions for defined fluents and actions, observations, actions, and defaults from $\mathcal{H}$, and a CR rule for every default allowing the agent to assume that the default's conclusion is false in order to restore consistency under exceptional circumstances. For example, the CR-Prolog statement:
\begin{align*}
  \lnot spread\_attack(X) \rif attacker(X)
\end{align*}
allows the ad hoc agent to consider the rare situation of attackers mounting a frontal attack.  Furthermore, it includes helper axioms, \emph{e.g.,} to define goals and drive diagnosis. Reasoning tasks such as planning, diagnosis, and inference are then reduced to computing answer sets of $\Pi$. The ad hoc agent may need to prioritize different goals at different times, \emph{e.g.,} score a goal when it has control of the ball, and position itself at a suitable location otherwise:
\begin{align}
    &goal(I) \leftarrow holds(scored\_goal,I) \\\nonumber 
    &goal(I) \leftarrow holds(loc(ad\_hoc\_agent, X, Y), I).
\end{align}
A suitable goal is selected and included in $\Pi(\mathcal{D}, \mathcal{H})$ automatically at run-time. In addition, heuristics are encoded to direct the search for plans, \textit{e.g.,} the following statements: 
\begin{align}
    &total(S) \leftarrow S = sum\{C,A:occurs(A,I), cost(A,C)\}\\\nonumber 
    &\#minimize \{S@p, S:total(S)\}.
\end{align}
encourage the ad hoc agent to select actions that will minimize the total cost when computing action sequences to achieve a particular goal. We use the SPARC system~\cite{balai:lpnmr13} to write and solve CR-Prolog programs. For computational efficiency, these examples programs build on a refinement-based architecture that represents and reasons with knowledge at two tightly-coupled resolutions, with a fine-resolution description ($\mathcal{D}_F$) defined as a refinement of a coarse-resolution description ($\mathcal{D}_C$). For example, the available space in the FA domain and HFO domain is organized into abstract regions in $\mathcal{D}_C$, with each region being refined in $\mathcal{D}_F$ into small grids that are components of this region. $\mathcal{D}_C$ and $\mathcal{D}_F$ are formally coupled through component relations and bridge axioms such as:
\begin{align*}
    &loc^*(A, Rg) ~~\mathbf{if}~~ loc(R, X, Y), ~component(X, Y, Rg)\\
    &next\_to^*(Rg_2, Rg_1) ~~\mathbf{if}~~ next\_to^*(Rg_1, Rg_2)
\end{align*}
where location $(X, Y)$ is in region $Rg$ and superscript ``*" refers to relations in $\mathcal{D}_C$. This coupling between the descriptions enables the ad hoc agent to automatically choose the relevant part of the relevant description based on the goal or abstract action, and to transfer relevant information between the descriptions. Example programs are in our repository~\cite{code-results-HFO}; details of the refinement-based architecture are in the paper by~\citeN{mohan:JAIR19}.

\subsection{Agent Models and Model Selection}
\label{sec:arch-models}
Since reasoning with just prior domain knowledge can lead to poor team performance under AHT settings (see Section~\ref{sec:expres-results}), KAT enables the ad hoc agent to also reason with models that predict (\emph{i.e.,} anticipate) the action choices of other agents. State of the methods attempt to optimize performance in different (known or potential) situations by learning models offline from many (\emph{e.g.,} hundred thousands or millions of) examples. It is intractable to obtain such labeled examples of different situations in complex domains, and the learned models are truly useful only if they can be learned and revised rapidly during run-time to account for previously unknown situations. KAT thus chooses relevant attributes for models that can be: (a) learned from limited (\emph{e.g.,} 10000) training examples acquired from simple hand-crafted policies (\emph{e.g.,} spread and shoot in FA, pass when possible in HFO); and (b) revised rapidly during run-time to provide reasonable accuracy. Tables~\ref{tab:attributes_FA} and~\ref{tab:attributes_HFO_team} list the identified attributes in the FA and HFO domain respectively. 

\begin{table}[tb]
\caption{Attributes considered for models of other agents' behavior in FA domain. Number of attributes represent the \emph{number of variables} in each attribute times the \emph{number of agents}.}
\label{tab:attributes_FA}
    \begin{minipage}[b]{0.48\textwidth}
    {\begin{tabular}{lr}
        \hline \hline
        Description of attribute  & Number \\
        \hline
        x, y position of agent & 12 \\ 
        distance from agent to center of field & 6 \\
        agents' polar angle with center of field & 6 \\
        orientation of the agent & 6 \\
        \hline \hline
    \end{tabular}}
    \end{minipage}
    \hfill
    \begin{minipage}[b]{0.48\textwidth}
    {\begin{tabular}{lr}
        \hline \hline
        Description of attribute  & Number \\
        \hline
        distance from agent to fort & 6 \\
        distance to nearest attacker from fort & 1 \\
        number of attackers not alive & 1 \\
        previous action of the agent & 1 \\
        \hline \hline
    \end{tabular}}
    \end{minipage}
\end{table}

\begin{table}[tb]
\caption{Attributes for models of teammates and defense agents' behavior in HFO domain. Number of attributes represent the \emph{number of variables} in each attribute times the \emph{number of agents}.}
\label{tab:attributes_HFO_team}
    \begin{minipage}[b]{0.48\textwidth}
    {\begin{tabular}{lr}
        \hline \hline
        Description of attribute  & Number \\
        \hline
        x position of agent & 4 \\ 
        y position of agent & 4 \\ 
        goal opening angle & 2 \\
        proximity to the nearest opponent & 2 \\
        x position of the ball & 1 \\
        y position of the ball & 1 \\
        \hline \hline
    \end{tabular}}
    \end{minipage}
    \hfill
    \begin{minipage}[b]{0.48\textwidth}
    {\begin{tabular}{p{4cm}r}
        \hline \hline
        Description of attribute & Number \\
        \hline
        x position of agent & 4 \\ 
        y position of agent & 4 \\ 
        x position of the ball & 1 \\
        y position of the ball & 1 \\
        \hline \hline
    \end{tabular}}
    \end{minipage}
    \vspace{-1\baselineskip}
\end{table}

\begin{figure}[t]
\begin{center}
\begin{subfigure}{\columnwidth}
  \centering
  \includegraphics[width=0.9\columnwidth]{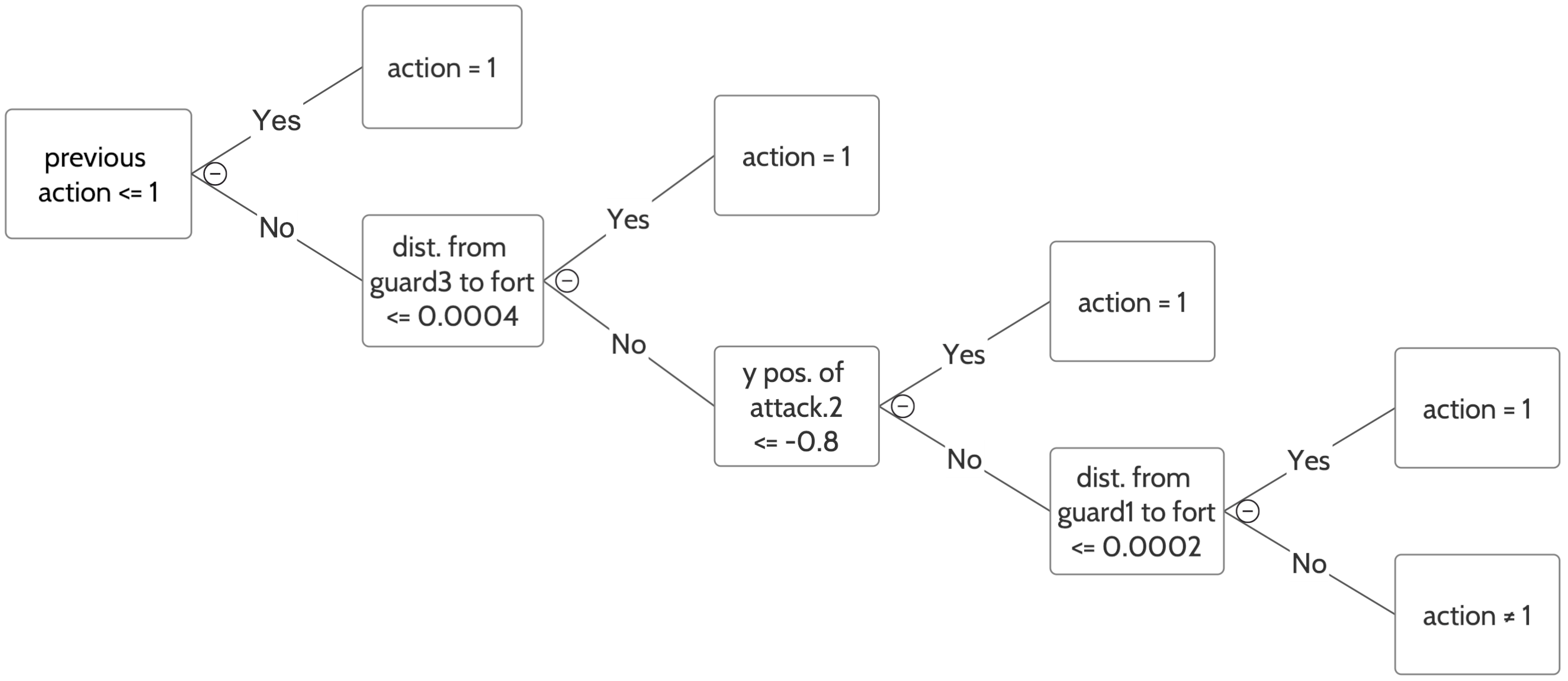}
  \caption{One FF tree in the ensemble for a guard in the FA domain.}
  \label{fig:tree-guard}
\end{subfigure}  
\begin{subfigure}{\columnwidth}
  \centering
  \includegraphics[width=0.8\columnwidth]{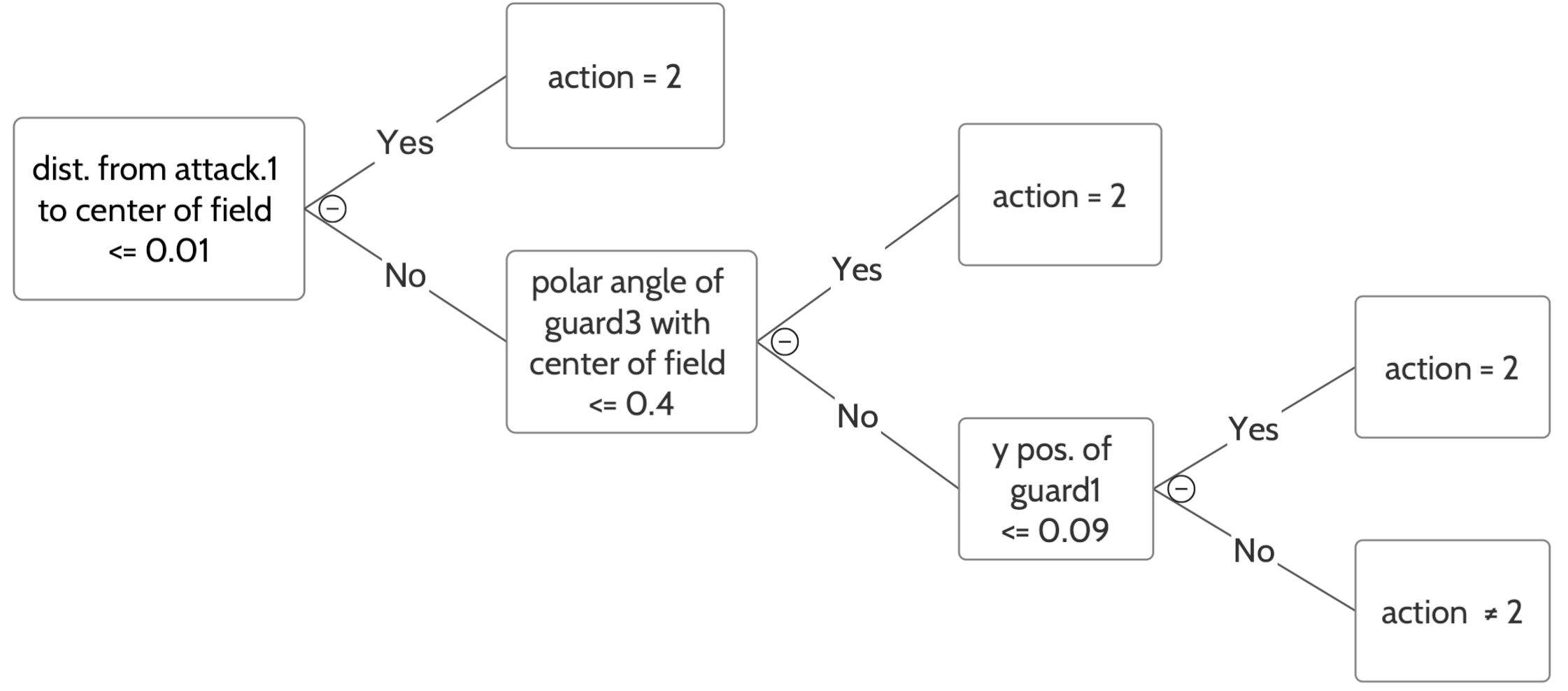}
  \caption{One FF tree in the ensemble for an attacker in the FA domain.}
  \label{fig:tree-attacker}
\end{subfigure}
\caption{Examples of FF trees for a guard and an attacker in the FA domain.}
\label{fig:tree-examples}
\end{center}
\end{figure}

Similar to our recent work~\cite{dodampegama:aaai23}, the attributes are identified and the predictive models are learned using the \textit{Ecological Rationality} (ER) approach, which draws on insights from human cognition, Herb Simon's definition of \textit{Bounded Rationality}, and an algorithmic model of heuristics~\cite{gigerenzer:bookchap20,gigerenzer:ARP11}. ER focuses on decision making under true uncertainty (\emph{e.g.,} in open worlds), characterizes behavior as a joint function of internal (cognitive) processes and the environment, and focuses on \emph{satisficing} based on differences between observed and predicted behavior. Also, heuristic methods are viewed as a strategy to ignore part of the information in order to make decisions more quickly, frugally, and/or accurately than complex methods. In addition, it advocates the use of an adaptive toolbox of classes of heuristics (e.g., one-reason, sequential search, lexicographic), and comparative out-of-sample testing to identify heuristics that best leverage the target domain's structure. This approach has provided good performance in many applications~\cite{gigerenzer:MMM16}. 

Specifically, in KAT, ER principles such as abstraction and refinement, and statistical attribute selection methods, are applied to the set of 10000 samples to identify the key attributes and their representation in Tables~\ref{tab:attributes_FA} and~\ref{tab:attributes_HFO_team}; these define behavior in the FA domain and HFO domain respectively. The coarse- and fine-resolution representation described in Section~\ref{sec:arch-krr} is an example of the principle of refinement. In addition to the choice of features, the characteristic factors of AHT, \textit{e.g.,} the need to make rapid decisions under resource constraints and respond to dynamic changes with limited examples, are matched with the toolbox of heuristics to identify and use an ensemble of ``fast and frugal" (FF) decision trees to learn the behavior prediction models for each type of agent. Each FF tree in an ensemble focuses on one valid action, provides a binary class label, and has the number of leaves limited by the number of attributes~\cite{katsikopoulos:book21}. Figure~\ref{fig:tree-examples} shows an example of a FF tree learned (as part of the corresponding ensemble) for a guard agent (Figure~\ref{fig:tree-guard}) and an attacker agent (Figure~\ref{fig:tree-attacker}) in the FA domain.


The ad hoc agent's teammates and opponents may include different types of agents whose behavior may change over time. \textit{Unlike our prior work that used static prediction models, we enabled the ad hoc agent to respond to such changes by automatically revising the current model, switching to a relevant model, or learning new models}. Existing models are revised by changing the parameters of the FF trees, and Algorithm~\ref{alg:model-selection} is an example of our approach for selecting a suitable model in the context of predicting the pose (\emph{i.e.,} position and orientation) of agents. Specifically, the ad hoc agent periodically compares the existing models' predictions with the observed action choices of each agent (teammate, opponent) over a sliding window of domain state and the agents' action choices; in Algorithm~\ref{alg:model-selection}, this window is of size 1 (Lines 4-5). Also, a graded strategy is used to compute the error, penalizing differences in orientation less than differences in position (Lines 6-7). The model whose predictions best match the observations is selected for subsequent use and revision (Line 10, Algorithm~\ref{alg:model-selection}). Note that if none of the models provide a good match over multiple steps, this acts as a trigger to learn a new model.

\begin{algorithm}[tb]
\normalsize
\caption{\textbf{Model Selection}}
\label{alg:model-selection}
\KwIn{$\mathcal{A}$: other agents; $\mathcal{M}$: subset of behavior models; $\{a_{act}\}, \{a_{pred}\}$: actual and predicted action choices of each agent in current round of the game; $scores$: initial values (100) assigned to each agent-model combination.}
\KwOut{model: selected model for each agent.}
\For{$i = 0$ \KwTo $\mathcal{A}$} {
    \For{$m = 0$ \KwTo $\mathcal{M}$}{
        \If{${a}_{pred}[i, m] \neq {a}_{act}[i]$}{
            ${l}_{act}, {o}_{act} \gets $ actual\_pose(${a}_{act}$)
            
            ${l}_{pred}, {o}_{pred} \gets $ predicted\_pose(${a}_{pred}$)

            $penalty \gets abs({l}_{act} - {l}_{pred}) + abs({o}_{act} - {o}_{pred})/10$
            
            $scores[i, m] = scores[i, m] - penalty$
        }
    } 
    model[i] = select\_model($\mathcal{M}$, scores[i, *])
} 
\end{algorithm}

\subsection{Partial Observability and Communication}
\label{sec:arch-partial}
In practical AHT domains, any single agent cannot observe the entire domain and communication is a scarce resource. To explore the interplay between partial observability and communication, we modified the original domains. Specifically, in the FA domain, we introduced a \textit{forest} region where attackers can hide from the view of the two guards other than ad hoc agent and secretly approach the fort---see Figure~\ref{fig:fa-l}. The ad hoc agent has visibility of the forest region; it can decide when to communicate with its teammates, \emph{e.g.,} when: (a) one or more attackers are hidden in the forest; and (b) one of the other guards is closer to the hidden attacker(s) than it. The associated reasoning can be encoded using statements such as:
\begin{subequations}
    \label{eqn:comm}
    \begin{align}
        holds(shoots(G, AA),I+1) \leftarrow &occurs(communicate(AHA, G, AA),I) \\
        holds(in\_forest(AA),I) \leftarrow &holds(agent\_loc(AA,X,Y),I),~forest(X,Y),\\ \nonumber
        &not~~holds(shot(AA),I) \\
        -occurs(communicate(AHA&, G, AA),I) \leftarrow ~not~holds(in\_range(G, AA),I).
    \end{align}
\end{subequations}
where Statement~\ref{eqn:comm}(c) encodes that communication is used only when a hidden attacker is within the range of a teammate (\textit{i.e.,} guard agent); Statement~\ref{eqn:comm}(b) defines when an attacker is hidden; and Statement~\ref{eqn:comm}(a) describes the ad hoc agent's belief that a teammate receiving information about a hidden attacker will shoot it, although the teammate acts independently and may choose to ignore this information. If there are multiple guards satisfying these conditions, the ad hoc agent may only communicate with the guard closest to the hidden attacker(s). 

In the HFO domain, we represent partial observability in an indirect manner using the built-in ability to limit each agent's perception to a specific viewing cone relative to the agent. Specifically, each agent is only able to sense objects (\emph{e.g.,} other agents, ball) within its viewing cone; objects outside its viewing cone are not visible. Given this use of built-in functions, we added some helper axioms to ensure that the ad hoc agent only reasoned with visible objects; no additional communication action was implemented.

\section{Experimental setup and results}
\label{sec:expres}
We experimentally evaluated three hypotheses about KAT's capabilities: 
\begin{itemize}
    \item[\textbf{H1:}] KAT's performance is comparable or better than state of the art baselines in different scenarios while requiring much less training;
    \item[\textbf{H2:}] KAT enables adaptation to unforeseen changes in the type and number of other agents (teammates and opponents); and
    \item[\textbf{H3:}] KAT supports adaptation to partial observability with limited communication capabilities.
\end{itemize}
We evaluated aspects of H1 and H2 in both domains (FA, HFO) under full observability. For H3, we considered partial observability in both domains, and explored limited communication in the FA domain. Each game (\emph{i.e.,} episode) in the FA domain had three guards and three attackers, with our ad hoc agent replacing one of the guards. In HFO domain, each game (\emph{i.e.,} episode) had two offense and two defense players (including one goalkeeper) in the limited version; and four offense and five defense players (including one goalkeeper) in the full version. Our ad hoc agent replaced one of the offense agents in the HFO domain. In the FA domain, the key performance measure was the win percentage of the guards team. In the HFO domain, the key performance measure was the fraction of games in which the offense team scored a goal. In both domains, we also measured the accuracy of the predictive models. Further details of the experiments and the associated baselines are provided below.

\subsection{Experimental Setup}
\label{sec:expres-setup}
In the \textbf{FA domain}, we used two kinds of policies for the agents other than our ad hoc agent: \textit{hand-crafted policies} and \textit{built-in policies}. Hand-crafted policies were constructed as simple strategies that produce basic behavior. Built-in policies were provided with the domain; they are based on graph neural networks trained using many labeled examples. 

\smallskip
\noindent
\textbf{Hand-Crafted Policies.}
\begin{itemize}
    \item \textbf{Policy1:} guards stay near the fort and shoot attackers who spread and approach.
    \item \textbf{Policy2:} guards and attackers spread and shoot their opponents.
\end{itemize}
\noindent
\textbf{Built-in Policies.}
\begin{itemize}
    \item \textbf{Policy220:} guards stay in front of the fort and shoot continuously as attackers approach.
    \item \textbf{Policy650:} guards try to block the fort; attackers try to sneak in from all sides.
    \item \textbf{Policy1240:} guards spread and shoot the attackers; attackers sneak in from all sides.
    \item \textbf{Policy1600:} guards are willing to move away from the fort; some attackers approach the fort and shoot to distract the guards while others try to sneak in.
\end{itemize}
The ad hoc agent was evaluated in two experiments: \textbf{Exp1}, in which other agents followed the hand-crafted policies; and \textbf{Exp2}, in which other agents followed the built-in policies. As stated earlier, the ad hoc agent learned behavior models in the form of FF trees from 10000 state-action observations obtained by running the hand-crafted policies. 
It was not provided any prior experience or models of the built-in policies. 

Our previous work documented the accuracy of a basic AHT architecture that reasoned with some domain knowledge and static behavior prediction models in the FA domain~\cite{dodampegama:aaai23}. In this paper, the focus is on evaluating the ability to select, revise, and learn the relevant predictive models, and adapt to partial observability. For the former, each agent other than our ad hoc agent was assigned a policy selected randomly from the available policies (described above). The baselines for this experiment were:
 \begin{itemize}
    \item \textbf{Base1:} other agents followed a random mix of hand-crafted policies. The ad hoc agent did not revise the learned behavior models or use the model selection algorithm.
    \item \textbf{Base2:} other agents followed a random mix of hand-crafted policies. The ad hoc agent used a model selection algorithm without a graded strategy to compare the predicted and actual actions, \textit{i.e.,} fixed penalty assigned for action mismatch in Line 6 of Algorithm~\ref{alg:model-selection}. 
    \item \textbf{Base3:} other agents followed a random mix of built-in policies. The ad hoc agent did not revise the learned behavior models or use the model selection algorithm. 
    \item \textbf{Base4:} other agents followed a random mix of built-in policies. The ad hoc agent used the model selection algorithm without a graded strategy to compare predicted and actual actions, \textit{i.e.,} fixed penalty assigned for action mismatch in Line 6 of Algorithm~\ref{alg:model-selection}.
\end{itemize}
The baselines for evaluating partial observability and communication were:
\begin{itemize}
    \item \textbf{Base5:} in \textbf{Exp1}, other agents followed hand-crafted policies and ad hoc agent did not use any communication actions.
    \item \textbf{Base6:} in \textbf{Exp2}, other agents followed built-in policies and the ad hoc agent did not use any communication actions.
\end{itemize}
Recall that KAT allows the use of communication actions (when needed) under conditions or partial observability. Also, each experiment described above (in FA domain) involved 150 episodes and results were tested for statistical significance.

\medskip
\noindent
In the \textbf{HFO domain}, we used six external agent teams from the 2013 RoboCup simulation competition to create the ad hoc agent's teammates and opponents. Five teams were used to create offense agents: \textit{helios, gliders, cyrus, axiom and aut}; agents of the defense team were based on the \textit{agent2d} team. Similar to the initial phase in the FA domain, we deployed the existing agent teams in the HFO domain and collected observations of states before and after each transition in the episode. Since the actions of other agents are not directly observable, they were computed from the observed state transitions. To evaluate the ability to learn from limited data, we only used data from 300 episodes for each type of agent to create the tree-based models for behavior prediction, which were then revised (as needed) and used by the ad hoc agent during reasoning.

We first compared KAT's performance with a baseline that only used non-monotonic logical reasoning with prior knowledge but without any behavior prediction models (\textbf{Exp3}), \emph{i.e.,} the ad hoc agent was unable to anticipate the actions of other agents. Next, we evaluated KAT's performance with each built-in external team, \emph{i.e.,} all offense agents other than the ad hoc agent were based on one randomly selected external team in each episode. In \textbf{Exp4}, we measured performance in the limited version, \emph{i.e.,} two offense players (including ad hoc agent) against two defense agents (including goalkeeper). In \textbf{Exp5}, we measured performance in the full version, \emph{i.e.,} four offense players (including ad hoc agent) played against five defense agents (including goalkeeper). In \textbf{Exp6} and \textbf{Exp7}, we evaluated performance under partial observability in the limited and full versions respectively. 
As the baselines for \textbf{Exp4-Exp5}, we used recent (state of the art) AHT methods: PPAS~\cite{santos:bookchap21}, and PLASTIC~\cite{Barrett:AIJ17}. These methods considered the same external agent teams mentioned above, allowing us to compare our results with the results reported in their papers. For \textbf{Exp6-Exp7}, we used the external agent teams as baselines. We conducted 1000 episodes for each experiment described above, and tested the results for statistical significance. 



\subsection{Experiment Results}
\label{sec:expres-results}
We begin with the results of experiments in the \textbf{FA domain}. First, Table~\ref{tab:sw_wins_hcp} summarizes the results of using our model selection algorithm in \textbf{Exp1}. When the other agents followed the hand-crafted policies and the model selection mechanism was not used by the ad hoc agent (\textbf{Base1}), the team of guards had the lowest winning percentage. When the ad hoc agent used the model selection algorithm with a fixed penalty assigned for any mismatch between predicted and actual actions (\textbf{Base2}), the performance of the team of guards improved. When the ad hoc agent used KAT's model selection method (Algorithm~\ref{alg:model-selection}), the winning percentage of the team of guards was substantially higher than the other two options. These results demonstrated that \textit{KAT's adaptive selection of the behavior prediction models improved performance}.

\begin{table}[tb]
\begin{minipage}[b]{0.48\textwidth}
    \centering
    \caption{Wins (\%) for guards with hand-crafted policies in FA domain (\textbf{Exp1}). Model adaptation improves performance.}
    \label{tab:sw_wins_hcp}
    {\begin{tabular}{lr}
        \hline \hline
        Experiment  & Win \% \\
        \hline
        Without model selection (\textbf{Base1})    & 63  \\
        When using direct comparison (\textbf{Base2})   & 68  \\
        With model selection algorithm (\textbf{KAT})  & 73 \\
        \hline \hline
    \end{tabular}}
\end{minipage}
\hfill
\begin{minipage}[b]{0.48\textwidth}
    \centering
    \caption{Wins (\%) for guards with built-in policies in FA domain (\textbf{Exp2}). Model adaptation improves performance.} 
    \label{tab:sw_wins_bp}
    {\begin{tabular}{lr}
        \hline \hline
        Experiment  & Win \% \\
        \hline
        Without model selection (\textbf{Base3})     & 47  \\
        When using direct comparison (\textbf{Base4})    & 45  \\
        With model selection algorithm (\textbf{KAT})   & 55 \\
        \hline \hline
    \end{tabular}}
\end{minipage}
\end{table}

Next, the results of \textbf{Exp2} are summarized in Table~\ref{tab:sw_wins_bp}. We observed that KAT enabled the ad hoc agent to adapt to previously unseen teammates and opponents that used the FA domain's built-in policies, based on the model selection algorithm and the online revision of the behavior models learned from the hand-crafted policies. KAT provided the best performance compared with not using any model adaptation or selection (\textbf{Base3}), and when model selection assigned a fixed penalty for action mismatch (\textbf{Base4}). These results and Table~\ref{tab:sw_wins_hcp} support \textbf{H1} and \textbf{H2}.

\begin{table}[tb]
\begin{minipage}[b]{0.48\textwidth}
    \centering
    \caption{Wins (\%) for guards with hand-crafted policies in FA domain (\textbf{Exp1}). Communication addresses partial observability.} \label{tab:co_wins_hcp}
    {\begin{tabular}{cp{1.75cm}p{3cm}}
        \hline \hline
        Policy  & With Comm. (\%) & Without \newline Comm. (\%, \textbf{Base5}) \\
        \hline
        Policy1   & 73  & 58  \\
        Policy2   & 19  & 8   \\
        \hline \hline
    \end{tabular}}
\end{minipage}
\hfill
\begin{minipage}[b]{0.48\textwidth}
    \centering
    \caption{Wins (\%) for guards with built-in policies in FA domain (\textbf{Exp2}). Communication addresses partial observability.}
    \label{tab:co_wins_bp}
    {\begin{tabular}{lp{1.7cm}p{3cm}}
        \hline \hline
        Policy  & With Comm. (\%) & Without \newline Comm. (\%, \textbf{Base6})\\
        \hline
        Policy220   & 79  & 85  \\
        Policy650   & 42  & 41  \\
        Policy1240  & 46  & 43  \\
        Policy1600  & 18  & 17  \\
        \hline \hline
    \end{tabular}}
\end{minipage}
\end{table}

The results from \textbf{Exp1} under partial observability, with and without communication (\textbf{Base5}), are summarized in Table~\ref{tab:co_wins_hcp}. Recall that the other agents used the FA domain's hand-crafted policies in this experiment. When the communication actions were enabled for the ad hoc (guard) agent, the winning percentage of the team of guards was substantially higher than the winning percentage of the team of guards when they could not use the communication actions. \textbf{Policy2} was a particularly challenging scenario (because both guards and attackers can shoot), which justified the lower (overall) winning percentage.

Next, the results from \textbf{Exp2} under partial observability, with and without communication (\textbf{Base6}), are summarized in Table~\ref{tab:co_wins_bp}. Recall that the other agents used the FA domain's built-in policies. We observed that when the guards (other than the ad hoc agent) followed policies 650, 1240, or 1600, the winning percentage of the team of guards was comparable or higher when communication actions were enabled compared with when these actions were not available (\textbf{Base6}). With Policy 220, the performance of the team of guards was slightly worse when the communication actions were enabled. However, unlike the other policies, Policy 220 results in the guards spreading themselves in-front of the fort and shooting continuously. Under these circumstances, partial observability and communication strategies were not important factors in determining the outcome of the corresponding episodes. These results support hypothesis \textbf{H3}.

\medskip
\noindent
We next describe the results from the \textbf{HFO domain}. Table~\ref{tab:logical-reasoning-only} summarizes results of \textbf{Exp3}, which compared KAT's performance with a baseline that had the ad hoc agent only reasoning with prior knowledge, \textit{i.e.,} without any learned models predicting the behavior of other agents. With KAT, the fraction of goals scored by the offense team was significantly higher than with the baseline. These results emphasized the importance of learning and using the behavior prediction models, and indicated that leveraging the interplay between representation, reasoning, and learning leads to improved performance, which supports hypothesis \textbf{H1}. 

\begin{table}[t]
    \centering
    \caption{Fraction of goals scored (\textit{i.e.,} games won) by the offense team in HFO domain with and without the learned behavior prediction models (\textbf{Exp3}). Reasoning with prior domain knowledge but without the behavior prediction models has a negative impact on performance. } \label{tab:logical-reasoning-only}
    {\begin{tabular}{lrr}
        \hline \hline 
        Version &  KAT (\%) & Logical Reasoner (\%) \\
        \hline 
        Limited (2v2) & 79 & 67 \\
        Full (4v5) & 30 & 26 \\
        \hline \hline
    \end{tabular}}
\end{table}

\begin{table}[tb]
\begin{minipage}[b]{0.48\textwidth}
    \centering
    \caption{Prediction accuracy of the learned agent behavior models in limited (2v2) version of the HFO domain (\textbf{Exp4}).}\label{tab:model_accuracy_2v2}
    {\begin{tabular}{p{2cm}r}
         \hline \hline
         Agent Type & Accuracy (\%) \\
         \hline
         Helios     & 78.2 \\
         Gliders    & 83.2 \\
         Cyrus      & 69.5 \\
         Aut        & 72.4 \\
         Axiom      & 76.2 \\
         Agent2D    & 79.8 \\
        \hline \hline
    \end{tabular}}
\end{minipage}
\hfill
\begin{minipage}[b]{0.48\textwidth}
    \centering
    \caption{Prediction accuracy of the learned agent behavior models in full (4v5) version of the HFO domain (\textbf{Exp5}).}\label{tab:model_accuracy_4v5}
    {\begin{tabular}{p{2cm}r}
         \hline \hline 
         Agent Type & Accuracy (\%) \\
         \hline 
         Helios     & 86.0 \\
         Gliders    & 66.4 \\
         Cyrus      & 77.6 \\
         Aut        & 67.7 \\
         Axiom      & 73.6 \\
         Agent2D    & 71.9 \\
         \hline \hline
    \end{tabular}}
\end{minipage}
\end{table}

Next, the prediction accuracy of the learned behavior models created for the limited version (\textbf{Exp4}) and full version (\textbf{Exp5}) of the HFO domain are summarized in Tables~\ref{tab:model_accuracy_2v2} and~\ref{tab:model_accuracy_4v5} respectively. Recall that these behavior models were learned for the agents other than the ad hoc agent using data from 300 episodes (for each external agent type). This translated to orders of magnitude fewer training samples than the few hundred thousand training samples used by state of the art data-driven methods that do not reason with domain knowledge. The prediction accuracy varied over a range for the different agent types. Although the accuracy values were not very high, the models could be learned and revised quickly during run-time; also, these models resulted in good performance when the ad hoc agent also reasoned with prior knowledge. 

\begin{table}[tb]
    \centering
    \caption{Fraction of goals scored (\textit{i.e.,} games won) by the offense team in HFO domain in the limited version (2v2, \textbf{Exp4}) and full version (4v5, \textbf{Exp5}). KAT's performance comparable with the baselines in the limited version and much better than the baselines in the full version.}\label{tab:HFO_result}
    {\begin{tabular}{lrrr}
        \hline \hline 
        Version & KAT (\%) & PPAS (\%) & PLASTIC (\%) \\
        \hline 
        Limited (2v2) & 79 & 80 & 80 \\
        Full (4v5) & 30 & 20 & 20 \\
        \hline \hline
    \end{tabular}}
\end{table}

\begin{table}[tbh]
    \centering
    \caption{Goals scored (\textit{i.e.,} games won) by offense team in HFO domain under partial observability (\textbf{Exp6}, \textbf{Exp7}). KAT's performance comparable with baseline that had no ad hoc agents in the team but used training datasets that were orders of magnitude larger.} \label{tab:HFO_result_partial}
    {\begin{tabular}{lrr}
        \hline \hline 
        Version & KAT (\%) & Original Team (\%)\\
        \hline 
        Limited (2v2) & 71 & 76 \\
        Full (4v5) & 18 & 20 \\
        \hline \hline 
    \end{tabular}}
\end{table}

The results of \textbf{Exp4} and \textbf{Exp5} comparing KAT's performance with the state of the art baselines for the HFO domain (PPAS, PLASTIC) are summarized in Table~\ref{tab:HFO_result}. Recall that these data-driven baselines required orders of magnitude more training examples and did not support reasoning with prior domain knowledge. The fraction of goals scored (\textit{i.e.,} games won) by the team of offense agents including our ad hoc agent was comparable with the goals scored by the baselines for the limited version, and substantially better than goals scored by the baselines for the full version. These results strongly support hypotheses \textbf{H1} and \textbf{H2}.


The results of evaluating KAT under partial observability (in HFO domain) are summarized in Table~\ref{tab:HFO_result_partial} compared with teams of external agent types without any ad hoc agent. Although the results indicate that KAT's performance was slightly lower than the baseline teams without any ad hoc agents, the difference was not significant and mainly due to noise (\emph{e.g.,} in the perceived angle to the goal during certain episodes). The ability to provide performance comparable with teams whose training datasets were orders of magnitude larger strongly supports hypothesis \textbf{H3}.

In addition to the experimental results documented above, videos of experimental trials, including trials involving unexpected changes in the number and type of other agents, are provided in support of the hypotheses in our open-source repository~\cite{code-results-HFO}.

\section{Conclusions}
\label{sec:conclusions}
Ad hoc teamwork (AHT) refers to the problem of enabling an agent to collaborate with others without any prior coordination. This problem is representative of many practical multiagent collaboration applications. State of the art AHT methods are data-driven, requiring a large labeled dataset of prior observations to learn offline models that predict the behavior of other agents (or agent types) and determine the ad hoc agent's behavior. This paper described KAT, a knowledge-driven AHT architecture that supports non-monotonic logical reasoning with prior commonsense domain knowledge and predictive models of other agents' behaviors that are learned and revised rapidly online using heuristic methods. KAT leverages KR tools and the interplay between reasoning and learning to automate the online selection and revision of the behavior prediction models, and to guide collaboration and communication under partial observability and changes in team composition. Experimental results in two benchmark simulated domains, Fort Attack and Half Field Offense, demonstrated that KAT's performance is better than that of just the non-monotonic logical reasoning component, and is comparable or better than state of the art data-driven methods that require much larger training datasets, provide opaque models, and do not support rapid adaptation to previously unseen situations. 


Our architecture open up multiple directions for further research. For example, we will investigate the introduction of multiple ad hoc agents in the benchmark domains used in this paper and in other complex multiagent collaboration domains. We will also continue to explore the benefits of leveraging the interplay between reasoning and learning for AHT in teams of many more agents, including on physical robots collaborating with humans. In addition, we will build on other work in our group~\cite{mohan:JAAMAS23,mohan:ACS18} to demonstrate the ad hoc agent's ability to learn previously unknown domain knowledge. Furthermore, we will build on our recent work~\cite{dodampegama:aaai23} and the work of others in our group~\cite{mota:SNCS21} to enable the ad hoc agent to provide relational descriptions as explanations of its decisions and beliefs in response to different kinds of questions.  


\section*{Acknowledgments}
This work was supported in part by the U.S. Office of Naval Research Award N00014-20-1-2390. All conclusions are those of the authors alone.

\bibliographystyle{acmtrans}
\bibliography{new_tlp2egui}

\begin{thebibliography}{}

\bibitem[\protect\citeauthoryear{Balai, Gelfond, and Zhang}{Balai
  et~al\mbox{.}}{2013}]{balai:lpnmr13}
{\sc Balai, E.}, {\sc Gelfond, M.}, {\sc and} {\sc Zhang, Y.} 2013.
\newblock {Towards Answer Set Programming with Sorts}.
\newblock In {\em {International Conference on Logic Programming and
  Nonmonotonic Reasoning}}.

\bibitem[\protect\citeauthoryear{Balduccini and Gelfond}{Balduccini and
  Gelfond}{2003}]{balduccini:aaaisymp03}
{\sc Balduccini, M.} {\sc and} {\sc Gelfond, M.} 2003.
\newblock {Logic Programs with Consistency-Restoring Rules}.
\newblock In {\em AAAI Spring Symposium on Logical Formalization of Commonsense
  Reasoning}.

\bibitem[\protect\citeauthoryear{Baral, Gelfond, Pontelli, and Son}{Baral
  et~al\mbox{.}}{2022}]{Baral:AI22}
{\sc Baral, C.}, {\sc Gelfond, G.}, {\sc Pontelli, E.}, {\sc and} {\sc Son,
  T.~C.} 2022.
\newblock An action language for multi-agent domains.
\newblock {\em Artificial Intelligence\/}~{\em 302}, 103601.

\bibitem[\protect\citeauthoryear{Baral, Gelfond, Son, and Pontelli}{Baral
  et~al\mbox{.}}{2010}]{Baral:AAMAS10}
{\sc Baral, C.}, {\sc Gelfond, G.}, {\sc Son, T.~C.}, {\sc and} {\sc Pontelli,
  E.} 2010.
\newblock Using answer set programming to model multi-agent scenarios involving
  agents' knowledge about other's knowledge.
\newblock In {\em Proceedings of the International Joint Conference on
  Autonomous Agents and Multiagent Systems, AAMAS}. Vol.~1. 259–266.

\bibitem[\protect\citeauthoryear{Baral, Son, and Pontelli}{Baral
  et~al\mbox{.}}{2010}]{Baral:Springer10}
{\sc Baral, C.}, {\sc Son, T.~C.}, {\sc and} {\sc Pontelli, E.} 2010.
\newblock Reasoning about multi-agent domains using action language
  $\mathcal{C}$: A preliminary study.
\newblock In {\em Computational Logic in Multi-Agent Systems}, {J.~Dix},
  {M.~Fisher}, {and} {P.~Nov{\'a}k}, Eds. Springer Berlin Heidelberg, 46--63.

\bibitem[\protect\citeauthoryear{Barrett, Rosenfeld, Kraus, and Stone}{Barrett
  et~al\mbox{.}}{2017}]{Barrett:AIJ17}
{\sc Barrett, S.}, {\sc Rosenfeld, A.}, {\sc Kraus, S.}, {\sc and} {\sc Stone,
  P.} 2017.
\newblock Making friends on the fly: Cooperating with new teammates.
\newblock {\em Artificial Intelligence\/}~{\em 242}, 132--171.

\bibitem[\protect\citeauthoryear{Barrett, Stone, Kraus, and Rosenfeld}{Barrett
  et~al\mbox{.}}{2013}]{Barrett:AAAI13}
{\sc Barrett, S.}, {\sc Stone, P.}, {\sc Kraus, S.}, {\sc and} {\sc Rosenfeld,
  A.} 2013.
\newblock Teamwork with limited knowledge of teammates.
\newblock In {\em AAAI Conference on Artificial Intelligence}. Vol.~27.
  102--108.

\bibitem[\protect\citeauthoryear{Bowling and McCracken}{Bowling and
  McCracken}{2005}]{Bowling:AAAI05}
{\sc Bowling, M.} {\sc and} {\sc McCracken, P.} 2005.
\newblock Coordination and adaptation in impromptu teams.
\newblock In {\em National Conference on Artificial Intelligence}. 53–58.

\bibitem[\protect\citeauthoryear{Chen, Andrejczuk, Cao, and Zhang}{Chen
  et~al\mbox{.}}{2020}]{Chen:AAAI20}
{\sc Chen, S.}, {\sc Andrejczuk, E.}, {\sc Cao, Z.}, {\sc and} {\sc Zhang, J.}
  2020.
\newblock {AATEAM}: Achieving the ad hoc teamwork by employing the attention
  mechanism.
\newblock In {\em AAAI Conference on Artificial Intelligence}. 7095--7102.

\bibitem[\protect\citeauthoryear{Deka and Sycara}{Deka and
  Sycara}{2021}]{Deka:20}
{\sc Deka, A.} {\sc and} {\sc Sycara, K.} 2021.
\newblock Natural emergence of heterogeneous strategies in artificially
  intelligent competitive teams.
\newblock In {\em Advances in Swarm Intelligence}, {Y.~Tan} {and} {Y.~Shi},
  Eds. Springer International Publishing, Cham, 13--25.

\bibitem[\protect\citeauthoryear{Dodampegama and Sridharan}{Dodampegama and
  Sridharan}{2023a}]{dodampegama:aaai23}
{\sc Dodampegama, H.} {\sc and} {\sc Sridharan, M.} 2023a.
\newblock {Back to the Future: Toward a Hybrid Architecture for Ad Hoc
  Teamwork}.
\newblock In {\em AAAI Conference on Artificial Intelligence}.

\bibitem[\protect\citeauthoryear{Dodampegama and Sridharan}{Dodampegama and
  Sridharan}{2023b}]{code-results-HFO}
{\sc Dodampegama, H.} {\sc and} {\sc Sridharan, M.} 2023b.
\newblock {Code}.
\newblock \url{https://github.com/hharithaki/KAT}.

\bibitem[\protect\citeauthoryear{Gelfond and Inclezan}{Gelfond and
  Inclezan}{2013}]{gelfond:ANCL13}
{\sc Gelfond, M.} {\sc and} {\sc Inclezan, D.} 2013.
\newblock {Some Properties of System Descriptions of $AL_d$}.
\newblock {\em Applied Non-Classical Logics, Special Issue on Equilibrium Logic
  and ASP\/}~{\em 23,\/}~1-2, 105--120.

\bibitem[\protect\citeauthoryear{Gigerenzer}{Gigerenzer}{2016}]{gigerenzer:MMM16}
{\sc Gigerenzer, G.} 2016.
\newblock {\em {Towards a Rational Theory of Heuristics}}.
\newblock Palgrave Macmillan UK, London, 34--59.

\bibitem[\protect\citeauthoryear{Gigerenzer}{Gigerenzer}{2020}]{gigerenzer:bookchap20}
{\sc Gigerenzer, G.} {2020}.
\newblock {What is Bounded Rationality?}
\newblock In {\em {Routledge Handbook of Bounded Rationality}}. {Routledge}.

\bibitem[\protect\citeauthoryear{Gigerenzer and Gaissmaier}{Gigerenzer and
  Gaissmaier}{2011}]{gigerenzer:ARP11}
{\sc Gigerenzer, G.} {\sc and} {\sc Gaissmaier, W.} 2011.
\newblock {Heuristic Decision Making}.
\newblock {\em Annual Review of Psychology\/}~{\em 62}, 451--482.

\bibitem[\protect\citeauthoryear{Hausknecht, Mupparaju, Subramanian,
  Kalyanakrishnan, and Stone}{Hausknecht
  et~al\mbox{.}}{2016}]{Hausknecht:ALA16}
{\sc Hausknecht, M.}, {\sc Mupparaju, P.}, {\sc Subramanian, S.}, {\sc
  Kalyanakrishnan, S.}, {\sc and} {\sc Stone, P.} 2016.
\newblock Half field offense: An environment for multiagent learning and ad hoc
  teamwork.
\newblock In {\em AAMAS Adaptive Learning Agents Workshop}.

\bibitem[\protect\citeauthoryear{Katsikopoulos, Simsek, Buckmann, and
  Gigerenzer}{Katsikopoulos et~al\mbox{.}}{2021}]{katsikopoulos:book21}
{\sc Katsikopoulos, K.}, {\sc Simsek, O.}, {\sc Buckmann, M.}, {\sc and} {\sc
  Gigerenzer, G.} 2021.
\newblock {\em {Classification in the Wild: The Science and Art of Transparent
  Decision Making}}.
\newblock MIT Press.

\bibitem[\protect\citeauthoryear{Macke, Mirsky, and Stone}{Macke
  et~al\mbox{.}}{2021}]{Macke:AAAI21}
{\sc Macke, W.}, {\sc Mirsky, R.}, {\sc and} {\sc Stone, P.} 2021.
\newblock Expected value of communication for planning in ad hoc teamwork.
\newblock In {\em AAAI Conference on Artificial Intelligence}. 11290--11298.

\bibitem[\protect\citeauthoryear{Mirsky, Carlucho, Rahman, Fosong, Macke,
  Sridharan, Stone, and Albrecht}{Mirsky et~al\mbox{.}}{2022}]{mirsky:eumas22}
{\sc Mirsky, R.}, {\sc Carlucho, I.}, {\sc Rahman, A.}, {\sc Fosong, E.}, {\sc
  Macke, W.}, {\sc Sridharan, M.}, {\sc Stone, P.}, {\sc and} {\sc Albrecht,
  S.} 2022.
\newblock {A Survey of Ad Hoc Teamwork: Definitions, Methods, and Open
  Problems}.
\newblock In {\em {European Conference on Multiagent Systems}}.

\bibitem[\protect\citeauthoryear{Mota, Sridharan, and Leonardis}{Mota
  et~al\mbox{.}}{2021}]{mota:SNCS21}
{\sc Mota, T.}, {\sc Sridharan, M.}, {\sc and} {\sc Leonardis, A.} 2021.
\newblock {Integrated Commonsense Reasoning and Deep Learning for Transparent
  Decision Making in Robotics}.
\newblock {\em Springer Nature CS\/}~{\em 2,\/}~242.

\bibitem[\protect\citeauthoryear{Rahman, Hopner, Christianos, and
  Albrecht}{Rahman et~al\mbox{.}}{2021}]{Rahman:ICML21}
{\sc Rahman, M.~A.}, {\sc Hopner, N.}, {\sc Christianos, F.}, {\sc and} {\sc
  Albrecht, S.~V.} 2021.
\newblock Towards open ad hoc teamwork using graph-based policy learning.
\newblock In {\em International Conference on Machine Learning}. 8776--8786.

\bibitem[\protect\citeauthoryear{Santos, Ribeiro, Sardinha, and Melo}{Santos
  et~al\mbox{.}}{2021}]{santos:bookchap21}
{\sc Santos, P.~M.}, {\sc Ribeiro, J.~G.}, {\sc Sardinha, A.}, {\sc and} {\sc
  Melo, F.~S.} 2021.
\newblock Ad hoc teamwork in the presence of non-stationary teammates.
\newblock In {\em Progress in Artificial Intelligence}, {G.~Marreiros}, {F.~S.
  Melo}, {N.~Lau}, {H.~Lopes~Cardoso}, {and} {L.~P. Reis}, Eds. Springer
  International, 648--660.

\bibitem[\protect\citeauthoryear{Son and Balduccini}{Son and
  Balduccini}{2018}]{Son:KI18}
{\sc Son, T.} {\sc and} {\sc Balduccini, M.} 2018.
\newblock Answer set planning in single- and multi-agent environments.
\newblock {\em Künstliche Intelligenz\/}~{\em 32}.

\bibitem[\protect\citeauthoryear{Son, Pontelli, and Nguyen}{Son
  et~al\mbox{.}}{2010}]{Son:Springer10}
{\sc Son, T.~C.}, {\sc Pontelli, E.}, {\sc and} {\sc Nguyen, N.-H.} 2010.
\newblock Planning for multiagent using asp-prolog.
\newblock In {\em Computational Logic in Multi-Agent Systems}, {J.~Dix},
  {M.~Fisher}, {and} {P.~Nov{\'a}k}, Eds. Springer Berlin Heidelberg, 1--21.

\bibitem[\protect\citeauthoryear{Son and Sakama}{Son and
  Sakama}{2010}]{son:bookchap10}
{\sc Son, T.~C.} {\sc and} {\sc Sakama, C.} 2010.
\newblock {Reasoning and Planning with Cooperative Actions or Multiagents using
  Answer Set Programming}.
\newblock In {\em Declarative Agent Languages and Technologies VII}. {Lecture
  Notes in Computer Science}, vol. 5948. {Springer Berlin Heidelberg},
  208--227.

\bibitem[\protect\citeauthoryear{Sridharan, Gelfond, Zhang, and
  Wyatt}{Sridharan et~al\mbox{.}}{2019}]{mohan:JAIR19}
{\sc Sridharan, M.}, {\sc Gelfond, M.}, {\sc Zhang, S.}, {\sc and} {\sc Wyatt,
  J.} 2019.
\newblock {REBA: A Refinement-Based Architecture for Knowledge Representation
  and Reasoning in Robotics}.
\newblock {\em Journal of Artificial Intelligence Research\/}~{\em 65},
  87--180.

\bibitem[\protect\citeauthoryear{Sridharan and Meadows}{Sridharan and
  Meadows}{2018}]{mohan:ACS18}
{\sc Sridharan, M.} {\sc and} {\sc Meadows, B.} 2018.
\newblock {Knowledge Representation and Interactive Learning of Domain
  Knowledge for Human-Robot Collaboration}.
\newblock {\em Advances in Cognitive Systems\/}~{\em 7}, 77--96.

\bibitem[\protect\citeauthoryear{Sridharan and Mota}{Sridharan and
  Mota}{2023}]{mohan:JAAMAS23}
{\sc Sridharan, M.} {\sc and} {\sc Mota, T.} 2023.
\newblock {Towards Combining Commonsense Reasoning and Knowledge Acquisition to
  Guide Deep Learning}.
\newblock {\em {Autonomous Agents and Multi-Agent Systems}\/}~{\em 37,\/}~4.

\bibitem[\protect\citeauthoryear{Stone, Kaminka, Kraus, and Rosenschein}{Stone
  et~al\mbox{.}}{2010}]{Stone:AAAI10}
{\sc Stone, P.}, {\sc Kaminka, G.}, {\sc Kraus, S.}, {\sc and} {\sc
  Rosenschein, J.} 2010.
\newblock {Ad Hoc Autonomous Agent Teams: Collaboration without
  Pre-Coordination}.
\newblock In {\em AAAI Conference on Artificial Intelligence}. 1504--1509.

\bibitem[\protect\citeauthoryear{Wu, Zilberstein, and Chen}{Wu
  et~al\mbox{.}}{2011}]{Wu:ijcai11}
{\sc Wu, F.}, {\sc Zilberstein, S.}, {\sc and} {\sc Chen, X.} 2011.
\newblock Online planning for ad hoc autonomous agent teams.
\newblock In {\em International Joint Conference on Artificial Intelligence}.
  439–445.

\bibitem[\protect\citeauthoryear{Zand, Parker-Holder, and Roberts}{Zand
  et~al\mbox{.}}{2022}]{Zand:IFAAMAS22}
{\sc Zand, J.}, {\sc Parker-Holder, J.}, {\sc and} {\sc Roberts, S.~J.} 2022.
\newblock On-the-fly strategy adaptation for ad-hoc agent coordination.
\newblock In {\em International Conference on Autonomous Agents and Multiagent
  Systems}. 1771–1773.

\bibitem[\protect\citeauthoryear{Zintgraf, Devlin, Ciosek, Whiteson, and
  Hofmann}{Zintgraf et~al\mbox{.}}{2021}]{zintgraf:21}
{\sc Zintgraf, L.}, {\sc Devlin, S.}, {\sc Ciosek, K.}, {\sc Whiteson, S.},
  {\sc and} {\sc Hofmann, K.} 2021.
\newblock Deep interactive bayesian reinforcement learning via meta-learning.
\newblock In {\em International Conference on Autonomous Agents and Multiagent
  Systems}.

\end{thebibliography}

\label{lastpage}


\end{document}